\documentclass[10pt,twocolumn,letterpaper]{article}

\usepackage[pagenumbers]{cvpr} 

\newcommand{\mysubsubsection}[1]{\vspace{0.1cm} \noindent {\bf #1}.}

\newcommand\mypara[1]{\vspace{0.3mm}\noindent\textbf{#1.}}

\newcommand\myparaa[1]{\vspace{0.3mm}\noindent\textbf{#1}}

\definecolor{cvprblue}{rgb}{0.21,0.49,0.74}
\usepackage[pagebackref,breaklinks,colorlinks,allcolors=cvprblue]{hyperref}

\def\paperID{***} 
\def\confName{CVPR}
\def\confYear{2026}

\title{B-Rep Distance Functions (BR-DF)\\How to Represent a B-Rep Model by Volumetric Distance Functions?}

\author{Fuyang Zhang$^1$ \qquad Pradeep Kumar Jayaraman$^2$ \qquad Xiang Xu$^2$ \qquad Yasutaka Furukawa$^{1,3}$ \\ [2pt]
\small{$^1$Simon Fraser University \qquad $^2$Autodesk \qquad $^3$Wayve}\\[2pt]
{\tt\small \{fuyangz, furukawa\}@sfu.ca \qquad \{xiang.xu, pradeep.kumar.jayaraman\}@autodesk.com}
}

\begin{document}
\twocolumn[{
\maketitle
\vspace{-1em}
\centerline{
\includegraphics[width=\linewidth]{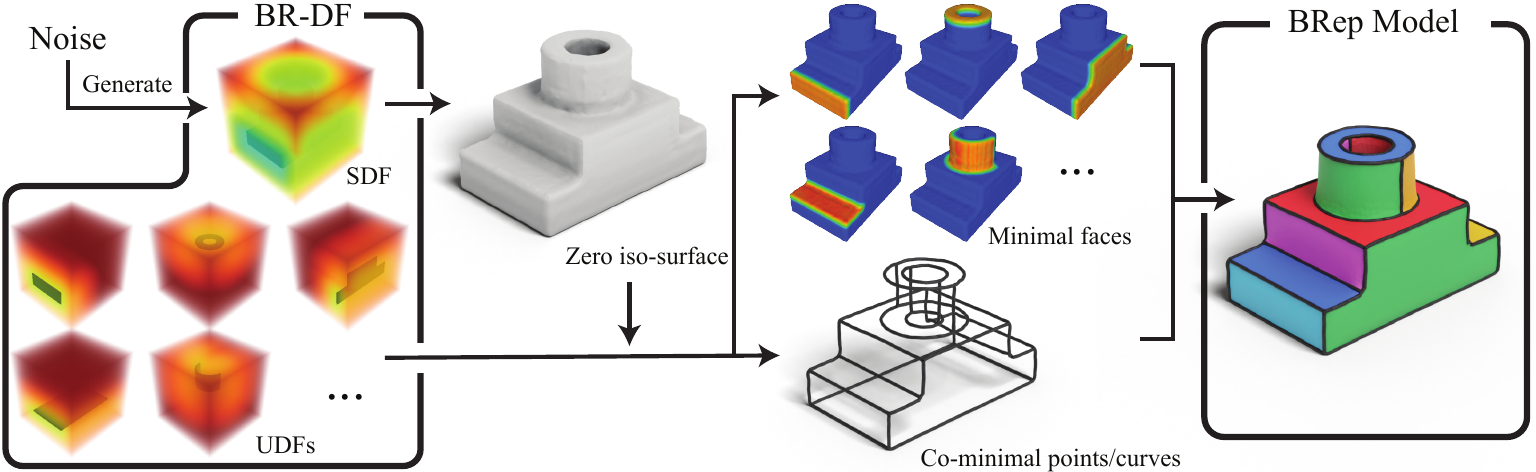}
}
\captionof{figure}{BR-DF is a geometric representation for Boundary Representation (B-Rep) models. An SDF encodes surface geometry. UDFs encode vertices, edges, faces, and their connectivity. An extension of the Marching Cubes  converts BR-DF to a faceted B-Rep model.} 
\label{fig:teaser}
\vspace{1em}
}]
\begin{abstract}
This paper presents a novel geometric representation for CAD Boundary Representation (B-Rep) based on volumetric distance functions, dubbed B-Rep Distance Functions (BR-DF). BR-DF encodes the surface mesh geometry of a CAD model as signed distance function (SDF). B-Rep vertices, edges, faces and their topology information are encoded as per-face unsigned distance functions (UDFs). 
An extension of the Marching Cubes algorithm converts BR-DF directly into watertight CAD B-Rep model (strictly speaking a faceted B-Rep model). A surprising characteristic of BR-DF is that this conversion process never fails.
Leveraging the volumetric nature of BR-DF, we propose a multi-branch latent diffusion with 3D U-Net backbone for jointly generating the SDF and per-face UDFs of a BR-DF model.
%
%
%
Our approach achieves comparable CAD generation performance against SOTA methods while reaching the unprecedented 100\% success rate in producing (faceted) B-Rep models.
\end{abstract}    
\section{Introduction}
\label{sec:intro}
Volumetric distance functions are fundamental in geometry processing. The introduction of the Marching Cubes algorithm~\cite{marchingcube} and signed distance functions (SDFs) marked a significant breakthrough in surface reconstruction for the field of computer graphics~\cite{curless1996volumetric,levoy2000digital}.
These concepts were later adopted by the computer vision community through level-set and geometric active contour methods~\cite{faugeras2002variational}, allowing geometry optimization without explicitly managing the surface topology changes.
In the era of deep learning and generative artificial intelligence (GenAI), signed distance functions remain central to geometry reconstruction and generation~\cite{dai2018scancomplete,park2019deepsdf,xu2019disn}, enabling high-quality 3D asserts creation through large-scale training~\cite{li2025triposg,zhang2024clay,hui2024make}. 

Solid modeling in computer-aided design (CAD) focuses on designing and editing shapes intended for real-world manufacturing. Ensuring watertightness is therefore a critical requirement and a key challenge for GenAI models designed to synthesize CAD data structures, such as Boundary Representations (B-Reps)~\cite{Weiler1986TopologicalSF}. Signed distance functions, by their very nature, guarantee watertight representations, making them a compelling choice for representing CAD B-Rep model in solid modeling. This paper advances the use of such volumetric distance functions by introducing an innovative representation that is both equivalent to and mutually convertible with a faceted B-Rep model~\cite{stroud2006boundary}, i.e., a segmented triangle mesh where each segment corresponds to a face in B-Rep, and the boundary curves between segments correspond to B-Rep edges.

Our proposed representation, B-Rep Distance Functions (BR-DF), combines a signed distance function (SDF) with per-face unsigned distance functions (UDFs) to fully capture the geometry and topology of a watertight CAD model. Surprisingly, we find that a simple algorithm can reliably convert BR-DF into faceted B-Rep---a process that \textit{never fails}. Specifically, we introduce a novel extension of the Marching Cubes algorithm that reconstructs the CAD surface boundary from SDF, and then conditioned on the surface mesh and UDFs, recovers the complete B-Rep vertices, edges, and faces together with their topology. 
Concretely, for a CAD model with $F$ faces, any point on the zero iso-surface of SDF has an associated $F$-dimensional face UDF scalar fields. A B-Rep face is determined by the minimal UDF value in the face UDF fields.
A B-Rep edge is defined when at least two faces share the same minimum UDF value, and similarly a B-Rep vertex occurs where at least three faces share the minimum UDF value.

Since our representation is a set of scalar functions on volumetric grids, we present a simple latent diffusion with 3D U-Net backbone for BR-DF generation.
Qualitative and quantitative evaluations on DeepCAD~\cite{deepcad} and ABC~\cite{abc} datasets demonstrate that BR-DF based generative model matches the performance of recent SOTA methods~\cite{deepcad, brepgen}, while achieving the unprecedented 100\% success rate of producing faceted B-Rep models.
While faceted B-Reps still needs to be converted into B-Reps using standard mesh-to-BRep conversion tools available in CAD software, the topology generated by our method greatly aids in this process.
We believe this representation will play a key role in developing generative models for CAD that guarantee watertightness.
To summarize, our paper makes three key contributions: 1) a novel volumetric distance function representation of B-Rep models, named BR-DF; 2) a Marching Cubes and Triangles algorithm to convert BR-DF to faceted B-rep models with 100\% success rate; 3) A simple B-Rep generative model based on BR-DF with performance competitive to the state-of-the-art approaches.
This paper potentially marks a key innovation in CAD geometry representation---analogous to the impact that Marching Cubes had on raster geometry modeling 25 years ago.

\section{Related Work}
This section reviews related work on distance functions and CAD model generation. 

\subsection{Signed Distance Functions}
Signed distance functions (SDFs) have been instrumental in geometry processing.
The introduction of the Marching Cubes algorithm~\cite{lorensen1998marching} enabled the creation of 3D mesh models from SDFs, setting the foundation for many surface reconstruction techniques. The Digital Michelangelo Project \cite{curless1996volumetric} is a notable example, which utilized SDFs to capture complex 3D geometry. In the Computer Vision domain, level-set and active contour methods utilized SDFs, allowing implicit surface representation that facilitated topology adaptation \cite{faugeras2002variational}. Over time, distance functions became less prominent in Computer Vision, as depth maps, point clouds, and mesh representations gained popularity; however, SDFs combined with Marching Cubes remain fundamental for surface extraction in geometry processing.

The integration of deep learning has renewed interest in distance functions.
Neural implicit representations, such as multilayer perceptron (MLP), predict SDF values for single-image reconstruction or scan completion~\cite{xu2019disn,park2019deepsdf,dai2018scancomplete}.
Volumetric grids are a popular alternative for the SDF representation and can be scaled to high resolutions~\cite{nanovdb,neuralwavelet}.
NeuralRecon reconstructs 3D scenes in real-time from a monocular video by predicting local TSDF volumes~\cite{sun2021neuralrecon}. Similarly, Atlas directly regresses a TSDF volume from a set of posed RGB images, utilizing a 2D convolutional neural network (CNN) to extract features from each image~\cite{murez2020atlas}.




%

\subsection{Generative Models for B-Reps}

Recent advancements in boundary representation (B-Rep) generation have focused on learning the construction sequences of sketch-and-extrude operations. This approach allows the generated sequences to be parsed by solid modeling kernels, enabling the rebuilding of B-Reps. DeepCAD~\cite{deepcad} pioneered this direction by releasing a dataset of sketch-and-extrude sequences, laying the groundwork for further research. Subsequent works introduced significant improvements: SkexGen~\cite{skexgen} leveraged autoregressive Transformer models~\cite{transformer} with disentangled learning of geometry and topology, leading to a substantial enhancement in generation quality. Later, HNC-CAD~\cite{hnccad} introduced hierarchical latent codes that capture natural hierarchies in the data structure, better preserving design intent and improving editability.
A key advantage of the sketch-and-extrude paradigm is its inherent guarantee of watertightness, as extruded 2D sketches are constrained to form closed loops. However, these methods rely on the availability of explicit construction sequences alongside B-Reps, limiting their applicability to simpler, primitive-based shapes.

Direct B-Rep generation creates 3D geometry while maintaining topological consistency among faces, edges, and vertices. SolidGen~\cite{jayaraman2022solidgen} employs Transformer models and pointer networks~\cite{pointernetworks} to generate B-Rep entities in a bottom-up manner. While this method benefits from larger datasets compared to sketch-extrude techniques, it remains constrained to basic geometric primitives and does not consume freeform data.
BrepGen~\cite{brepgen} introduced a structured latent representation, learning face and edge embeddings via uniformly sampled point grids in the parameter space~\cite{uvnet}. By incorporating diffusion models, it progressively generates faces and edges in a top-down manner, facilitating the creation of complex freeform surfaces while ensuring topological consistency. However, both of these methods adopt a surface-modeling approach to generating solid models and struggle with maintaining watertightness.
A particularly relevant method to our research is Split-and-Fit~\cite{splitandfit}, which reconstructs B-Reps from point clouds using learned Voronoi partitions. By integrating geometric algorithms and heuristics, it significantly improves watertightness in post-processing compared to prior approaches.
In contrast, our BR-DF representation offers a simpler and more robust alternative based on signed and unsigned distance functions, enabling conversion to faceted B-Reps using a straightforward algorithm with guaranteed success.

\section{B-Rep Distance Functions (BR-DF)}

\begin{figure*}[!t]
\includegraphics[width=\linewidth]{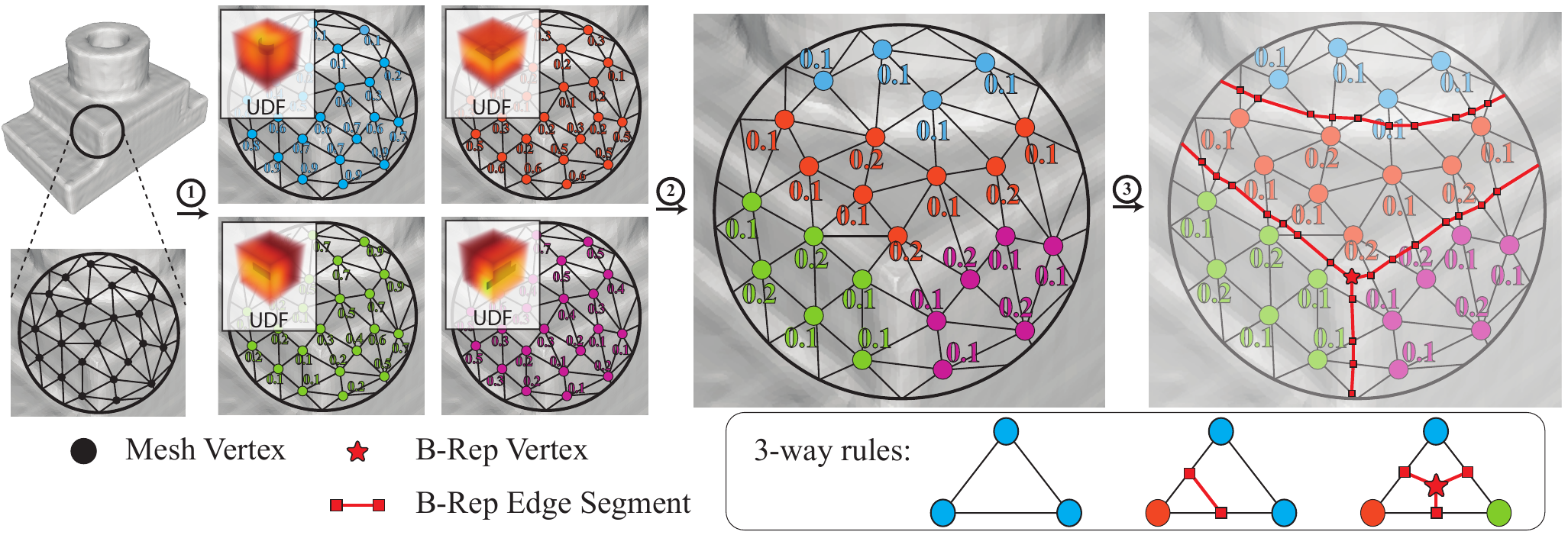}
\caption{MCT algorithm: 1) Given the volumetric UDFs for each face, interpolate a UDF value at each mesh vertex; 2) Select the face with the smallest UDF at each mesh vertex; and 3) Extract B-Rep vertices and edges by applying the 3-way rules.}
\label{fig:mcbt}
\end{figure*}

BR-DF is the core of our technical innovation. This section provides a detailed explanation of its representation.
Next section explains the novel extension of Marching Cubes algorithm that converts BR-DFs to faceted B-Rep models.


%

\subsection{Signed Distance Function (SDF)}
BR-DF consists of two volumetric distance functions (see~\autoref{fig:teaser}). The first function is the widely used signed distance function (SDF) that returns a signed distance to the nearest point on the surface, with negative values indicating inside and positive values for outside. 

\mysubsubsection{SDF zero iso-surface}
Following existing literature, the zero iso-surface of an SDF yields the surface geometry~\cite{curless1996volumetric,levoy2000digital} which we refer to as an ``SDF-surface.''
Note that our representation assumes a solid B-Rep model, which is watertight and has a well-defined SDF.

\subsection{Per-face Unsigned Distance Function (UDF)}
The second function is an unsigned distance function (UDF) that 
returns an unsigned distance to the closest point on the face of a CAD B-Rep model. When UDFs are sliced along the SDF zero iso-surface, they encode B-Rep vertices, edges, faces, and their connectivity as follows.

\mysubsubsection{UDF minimum face}
A straightforward interpretation of UDF would be to consider a set of points with the zero UDF value as a face on the SDF surface. However, UDFs are unsigned and never zero.
One might ask why not a per-face SDF,
with the zero iso-curve defining the face boundary. However, this leads to inconsistencies of the zero iso-curves between adjacent faces.
%
Instead, we consider a ``UDF minimum face'' as the set of points
where the UDF of a face is the minimum among all the faces. 
This definition effectively partitions the SDF-surface without overlaps or gaps.

\mysubsubsection{UDF co-minimum curve}
The UDF minimum face definition deduces ``UDF co-minimal curves".
B-Rep edges or a face boundary is a set of points where two face UDFs are the minimum simultaneously.



\mysubsubsection{UDF co-minimum point}
The UDF co-minimum curve definition deduces ``UDF co-minimum points". A B-Rep vertex is where three UDFs are the minimum simultaneously.
B-Rep points at X-junctions or higher-order degrees are degenerate configurations in the BR-DF representation, represented by two or more nearby co-minimum points.

\section{Marching Cubes and Triangles (MCT)} \label{sec:MCBT}



Marching Cubes and Triangles (MCT) takes a discrete BR-DF of distance functions sampled at 3D grid points as input and outputs
a faceted B-Rep model, i.e., a B-Rep model except for parametric curve/surface fitting. 
The remarkable aspect of the BR-DF representation and MCT algorithm is their robustness—this process never fails and consistently guarantees watertightness and valid topology. MCT always converts any BR-DF into a valid faceted B-Rep model.




\myparaa{Marching Cubes} extracts a triangulated mesh from a surface SDF. Triangle vertices are on the edges of the uniform grid, and per-face UDF values are linearly interpolated at the mesh vertices (see \autoref{fig:mcbt} Step 1 for details).

\myparaa{Marching Triangles} is a novel extension that extracts CAD B-Rep vertices, edges, faces, and their topology from the triangulated mesh by identifying UDF minimal faces and tracing the co-minimal points and curves. 
Marching Cubes applies a 256-way rule to each cube independently based on the signs of the SDF at its 8 grid points, achieving global consistency and a watertight mesh. Similarly, Marching Triangles applies a 3-way rule to each triangle independently depending on the face UDFs at its 3 vertices, achieving global consistency and a valid faceted B-Rep model.

Concretely, we determine the minimal face at each mesh vertex by selecting the face with the smallest UDF, applying perturbations in case of a tie (see Step 2 in \autoref{fig:mcbt}).
For each triangle, we extract a B-Rep vertex and/or edges via the following 3-way rule as illustrated in \autoref{fig:mcbt} Step 3: 1) if the minimal faces at all the vertices are the same, no vertex or edge are created; 2) if two vertices share the same minimal face, a B-Rep edge is created to separate the two vertices from the third by cutting edges where the two UDFs are equal via linear interpolation;
%
and 3) if all three minimal faces are different, a B-Rep vertex and three incident B-Rep edges are created based on the UDF values at these three vertices (detailed equations are in the supplementary).
Note that this operation is performed locally for each triangle but the algorithm does achieve global consistency.

\section{BR-DF Generative Model} \label{sec:method}
BR-DF is a set of 3D scalar fields with an arbitrary number of UDFs (i.e., faces).
To determine the number of UDFs, we follow BrepGen~\cite{brepgen}, which first generates axis-aligned bounding boxes of the faces.
Given the number of faces, a latent diffusion model with a 3D U-Net backbone generates the SDF and UDFs, using 3D VQ-VAEs as the latent encoder/decoder.
This section explains the bounding box generation, 3D VQ-VAEs, and the latent diffusion model.


\subsection{Preliminaries} \label{sec:preliminaries}

We represent an SDF or a UDF as $64^3$ values sampled at volumetric grid points within a cube. Given a B-Rep model, we determine its tightest axis-aligned bounding box and expand it to a cube while maintaining the center and adding a 20\% margin on all sides. We normalize the 3D coordinates so that the cube center is the origin, and its dimensions are 2.0. We employ simple heuristics to compute SDF and UDF values by using \textit{compute\_signed\_distance} and \textit{compute\_distance} from Open3D~\cite{zhou2018open3d}, and truncate results to $[-0.1, 0.1]$ range because important values are near zero.


\subsection{Bounding Box Generation}
We borrow the \textit{face position denoiser module} from BrepGen~\cite{brepgen}, which
predicts axis-aligned bounding boxes as the 3D coordinates of their bottom-left and top-right corners. It generates a predetermined maximum of 40 bounding boxes (60 during inference to minimize false negatives), while applying a non-maximum suppression (i.e., two bounding boxes become duplicates if the coordinates are within 0.04 distance for the two corners).




\subsection{3D VQ-VAEs} \label{sec:vq-vae}
We use a standard VQ-VAEs~\cite{vqvae} with 3D ResNet backbone~\cite{he2016deep} to
encode the distance function of $64^3$ spatial resolution into a latent volume of $4^3$ resolution.
The VQ-VAE architecture consists of four
%
downsampling and four upsampling blocks. Each block contains three layers of 3D-ResNet blocks. The feature dimensions for downsampling are 64-128-128-256, and in reverse order for upsampling.
We train two VQ-VAEs (one for an SDF and one for a UDF) using a combination of MSE reconstruction loss and codebook loss. The codebook size is set to $8192$.


\subsection{BR-DF Latent Diffusion Module}
\label{sec:latent_diffusion}
\begin{figure}[!t]
\centering
\includegraphics[width=\linewidth]{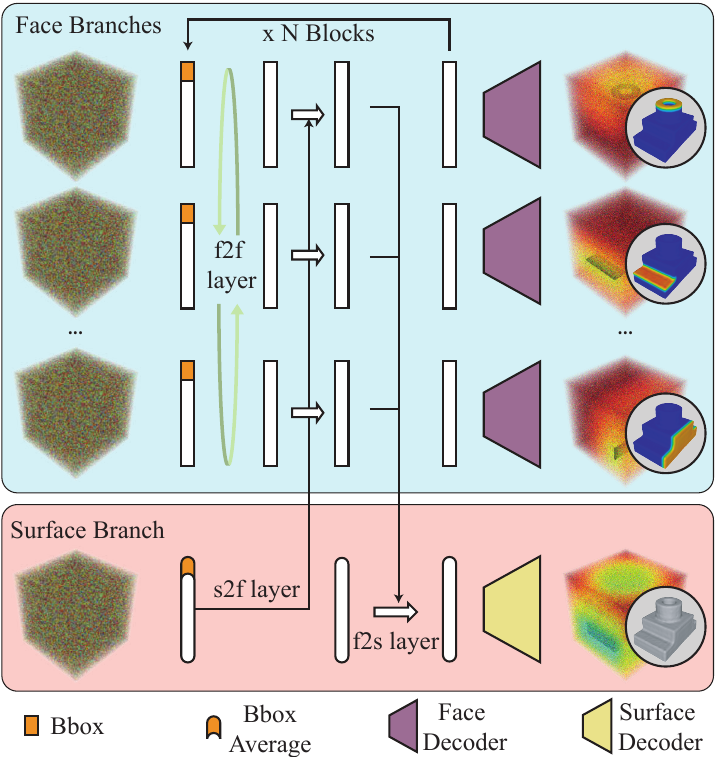}
\caption{The multi-branch diffusion architecture. Top and bottom represent the face and surface branches, respectively. The number of faces corresponds to the generated bounding boxes during inference and the ground truth bounding boxes during training. Three inter-branch cross-attention modules (f2f, f2s, s2f) are added.}
\label{fig:overview}
\end{figure}

BR-DF generation follows a multi-branch diffusion architecture~\cite{mvdiffusion}. It comprises of
a surface branch for the surface SDF and the face branches for the face UDFs (see~\autoref{fig:overview}). The number of face branches is set to the number of generated bounding boxes at inference and that of the ground-truth at training. 

The BR-DF Latent Diffusion Model Backbone is a 
3D U-Net~\cite{unet} with
three encoding and three decoding blocks. The first two encoding blocks are followed by a downsampling operation, \textit{nn.Conv3d(stride=2)}, and the last two decoding blocks have an upsampling operation, \textit{nn.Upsample(scale\_factor=2, model="nearest")}. Each block has two layers of 3D-ResNet blocks.
A two-layer MLP injects 6-DoF bounding box parameters into the face branch by adding it to the face feature. Similarly, the average of all bounding boxes is added to the surface branch. 

Three inter-branch layers are added following each U-Net block: face-to-face (f2f) layer, face-to-surface (f2s) layer, and surface-to-face (s2f) layer. 

\noindent $\bullet$
face-to-face (f2f) layer and face-to-surface (f2s) layer have the same structure. The source feature volume (a face volume for f2f layer and the surface volume for f2s layer) cross attention to the target feature volumes (remaining face features for f2f layer and all face features for f2s layer) at the same location. That means each token of the source feature attends only to the corresponding location in the target feature maps. A 3D ResNet block then refines the feature.

\noindent $\bullet$
surface-to-face (s2f) layer is a 3D ResNet block that takes an input as the concatenation of the surface volume and the face volume over the channel dimension, and outputs the updated face latent.

Loss Function. The model is trained to predict the L2-norm regression loss of the added sampled noise. 
\begin{equation}
\begin{aligned}
    L = \mathbb{E}_{t,\{Z_f^i\}_{i=1}^N,Z_s,{\{\epsilon^i\sim \mathcal{N}(0,I)\}}_{i=1}^{N},\epsilon\sim\mathcal{N}(0,I)} \\
    \left[ \sum_{i=1}^N
        \| \epsilon^i - \epsilon_f^i(\overline{Z^i_f}, t) \|^2 + \| \epsilon-\epsilon_s(\overline{Z_s}, t)\|^2
    \right],
\end{aligned}
\end{equation}
where $\epsilon, \epsilon^i$ are gaussian noises; $\epsilon_f,\epsilon_s$ denotes the model outputs; $Z_f^i$ and $Z_s$ are face latent and surface latent; $\overline{Z}$ is the latent with added noise.

\section{Experiments}
\label{sec:experiments}
This section presents the generation results of our approach. Extensive analysis demonstrates the advantages of B-Rep Distance Functions (BR-DF). Unlike traditional Boundary Representations which involves complex geometric structures, BR-DF employs volumetric-based distance functions that bypasses these complexities, making learning much easier. The Marching Cube and Triangles (MCT) algorithm ensures robust conversion to faceted B-Rep with 100\% watertightness and valid topology. The results show that our approach matches the state-of-the-art performance while having a clear advantage in terms of robustness.

\begin{figure*}[t]
    \centering
    \includegraphics[width=\linewidth]{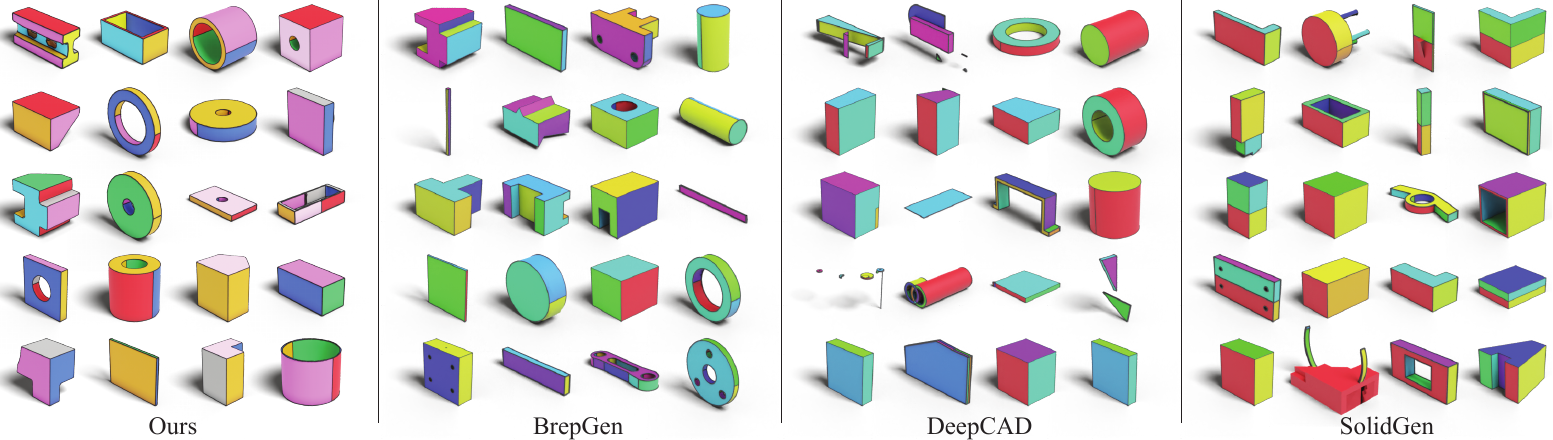} 
    \caption{Unconditional generation results on DeepCAD dataset~\cite{deepcad}. Our method achieves comparable performance to the state-of-the-art methods, while maintaining 100\% success rate.}
    \label{fig:comparison}
\end{figure*}

\subsection{Experiment Setup}

\mypara{Dataset} We evaluate generation quality on two datasets: 1) DeepCAD dataset~\cite{deepcad} comprises of mechanical parts made from sketch-and-extrude operations, 2) ABC dataset~\cite{abc} includes a wide variety of CAD parts in B-Rep formats from industrial design. Following~\cite{willis2021engineering}, we remove duplicated models in the datasets. Periodic surfaces (cylinders, etc.) are split on the seams following SolidGen and BrepGen~\cite{solidgen,brepgen}. B-Reps with more than 40 faces are also filtered out to manage memory usage. After filtering, a total of 55,897 DeepCAD data and 120,868 ABC data are used for training.

\mypara{Implementation Details} We implement our model in PyTorch, using AdamW~\cite{adamw} optimizer with a learning rate of with $5e^{-4}$ for all modules.
We use continuous latent for subsequent latent diffusion and the quantization layer is absorbed by the decoder.
Following BrepGen~\cite{brepgen}, the bounding box generation module is a standard Transformer diffusion model with pre-layer normalization, 12 self-attention layers, and 12 heads. Hidden dimension is 1024, and feature dimension is 768. Both bounding box generation module and BR-DF latent diffusion module use 1,000 diffusion steps and a linear beta schedule from $1e^{-4}$ to $0.02$. The bounding box generation module is trained for 500 epochs with a batch size 256 on a single NVIDIA A100 GPU, while the BR-DF latent diffusion module is trained for 1000 epochs with a batch size of 64 on two A100 GPUs. 
\subsection{Inference}
At inference, we use a 1,000 step DDPM~\cite{ddpm} to denoise the face bounding boxes. Following BrepGen, we increase the number of bounding box tokens to 60, which exceeds the maximum threshold used during training. This prevents noisy bounding boxes from incorrectly merging in the denoising process.
Non-maximum suppression eliminates duplicated bounding boxes. Next, the face UDFs and surface SDF in BR-DF are jointly denoised conditioned on the previously generated bounding boxes. We use a DDIM~\cite{ddim} scheduler with 200 steps for fast sampling. Finally, MCT extracts the faceted B-Rep from the generated BR-DF.


\subsection{Post-Processing}
Marching Cubes and Triangles (MCT) algorithm extracts faceted B-Rep models from BR-DF with 100\% success rate. To improve visual quality and support further B-Rep conversion,
we apply a conventional post-processing procedure to smooth both the mesh surface and its boundaries. We first compute parametric curves for each boundary while preserving T-junctions. We then perform multiple iterations of Laplacian smoothing and quadric-based edge-collapse remeshing to simplify and refine each face. During this process, boundary vertices at the same location from different faces are averaged to maintain geometric consistency. Please refer to the supplementary for further details.

{
\setlength{\tabcolsep}{3pt}
\begin{table}[t]
    \centering
    \begin{tabular}{lccc|ccc}
        \toprule
        Method & COV & MMD & JSD & Novel & Unique & Valid\\
        & $\%\uparrow$ & $\downarrow$ & $\downarrow$ & $\%\uparrow$ & $\%\uparrow$ & $\%\uparrow$\\
        \midrule
        DeepCAD  & 65.5 & 1.29 & 1.67 & 87.4 & 89.3 & 46.1 \\
        SolidGen & 71.0 & 1.08 & 1.31 & 99.1 & 96.2 & 60.3 \\
        BrepGen  & 73.9 & 1.04 & 1.28 & 99.8 & 99.7 & 62.9 \\
        Ours     & 73.7 & 1.09 & 1.40 & 99.9 & 99.1 & 100.*\\
        $\text{BrepGen}_{\text{ABC}}$ & 57.9 & 1.35 & 3.69 & 99.7 & 99.4 & 48.2 \\
        $\text{Ours}_\text{ABC}$ & 56.8 & 1.35 & 3.80 & 99.9 & 99.5 & 100.* \\
        \bottomrule
    \end{tabular}
    \caption{DeepCAD and ABC unconditional generation evaluation based on \textit{Coverage} (COV), \textit{Minimum Matching Distance} (MMD), \textit{Jensen-Shannon Divergence} (JSD) and \textit{Unique}, \textit{Novel}, \textit{Valid} ratio.}
    \label{table:quantitative}
\end{table}
}

\subsection{Evaluations}
Following BrepGen~\cite{brepgen}, we randomly sampled 3,000 results from the generated data and 1,000 from the reference test set to evaluate our model with two sets of metrics: distribution metrics and CAD metrics. The distribution metrics are based on 2,000 points sampled from the solid surface:
\begin{itemize}
    \item \textit{Coverage} (COV) is the ratio of reference data with at least one match after assigning each generation to its closest neighbor in the reference based on Chamfer Distance.
    \item \textit{Minimum Matching Distance} (MMD) is the average Chamfer Distance between a reference set and its nearest neighbor in the generated set.
    \item \textit{Jensen-Shannon Divergence} (JSD) measures the distribution distance between reference and generated data by converting points into $28^3$ discrete voxels.
\end{itemize}
CAD metrics compute the following scores:
\begin{itemize}
    \item \textit{Novel} percentage is the ratio of the generated data that does not appear in the training set.
    \item \textit{Unique} percentage is the ratio of the  data that appears only once in the generated results.
    \item \textit{Valid} percentage is the ratio of the generated B-Rep data that are watertight solids.
\end{itemize}

\begin{figure*}[t]
    \centering
    \includegraphics[width=\linewidth]{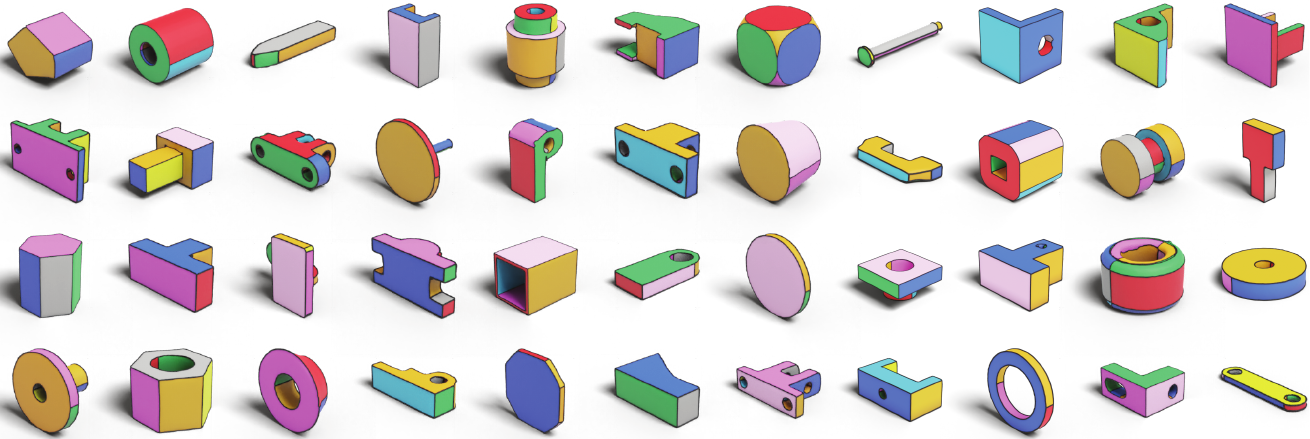} 
    \caption{Unconditional generation results on the ABC dataset~\cite{abc}.}
    \label{fig:abc_result}
\end{figure*}

\subsection{Unconditional B-Rep Generation}
We compare against DeepCAD~\cite{deepcad}, SolidGen~\cite{solidgen} and BrepGen~\cite{brepgen} for unconditional generation. For DeepCAD we further rebuild the generated sketch-and-extrude parameters into B-Rep solids. 

\mysubsubsection{Qualitative Evaluation}
Figure~\ref{fig:comparison} presents qualitative results on DeepCAD dataset. Our results demonstrate comparable performance to state-of-the-art methods. Additional results are provided in the supplementary materials. Figure~\ref{fig:abc_result} further showcases generated results trained on the ABC dataset. The key strength of our method is its unprecedented 100\% success rate in producing faceted B-Rep models. To the best of our knowledge, no existing method comes close to achieving this level of reliability.

\mysubsubsection{Quantitative Evaluation}
Table~\ref{table:quantitative} presents the quantitative results of our method against three other existing approaches. Our model is able to achieve comparable scores in COV, MMD, and JSD metrics. It is only slightly lower than the more complicated BrepGen method which has four different diffusion modules~\cite{brepgen}. Another reason is that our approach represents B-Rep through volumetric distance functions. This introduces a disadvantage for distribution-based metrics, as these metrics rely on sampling points from the mesh surface, making our performance sensitive to resolution. However, our key strength lies in achieving a 100\% validity rate. Validity is defined as generating watertight models with no topological errors—a criterion our method never fails. It is important to note that our success rate is calculated based on faceted B-Rep. Nonetheless, we argue that given a watertight and topologically valid structure, converting it into a standard B-Rep format is a straightforward and manageable process. Two B-Reps are considered identical if they have the same topological connections and if the 4-bit quantized positions of B-Rep vertices and edges (represented by the concatenation of their two endpoint positions) are also identical.

\subsection{Reconstruction Evaluation}
BR-DF representation is equivalent to a facated B-Rep model. We evaluate the numerical accuracy of this equivalence through the following steps: 1) Convert a faceted B-Rrep model into BR-DF via distance computation; 2) Reconstruct a faceted B-Rep model using MCT; and 3) Evaluate the reconstruction accuracy by comparing the reconstructed model against the original model. We use three metrics.
The first one is Chamfer Distance (CD) based on 5,000 points uniformly sampled on the surfaces of both models.
We compute CD metrics before and after the post-processing step.
The second one is Invalid Rate (IR) defined in DeepCAD~\cite{deepcad}, measuring the proportion of cases where B-Rep conversion failed due to invalid topology.
The third one is the Same Topology Rate (STR), which measures the ratio of samples where the CAD topology (i.e., the numbers of points, curves, and faces together with their connectivity) match exactly. Note that STR is more ``strict'' than IR, because the reconstruction must lead to a valid B-Rep model and the topology must match exactly.

In Table~\ref{table:recon_eval},  $\text{Ours}_\text{raw}$ shows the number for our  reconstruction procedure described above. $\text{Ours}_\text{latent}$ is a version where we further add the 3D VQ-VAE encoder and decoder (\autoref{sec:vq-vae}) to the BRDF representation. Lastly, we also evaluate DeepCAD system, where a transformer encodes a BRep-model into a 256 dimensional embedding and another transformer decodes an embedding back to a BRep-model.
The table shows that $\text{Ours}_\text{raw}$ achieves impressive 0\% IR and 100\% STR. CD is at the scale of $6.6e\text{-4}$, which is negligible given that the model coordinates are normalized to a scale of 2.0 (\autoref{sec:preliminaries}).




\subsection{Conditional B-Rep Generation}
\begin{figure}[t]
    \centering
    \includegraphics[width=0.9\linewidth]{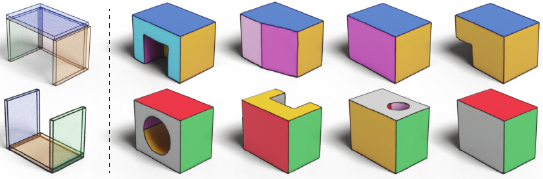} 
    \caption{Given partial bounding boxes (left), our model generates a diverse set of CAD models with complete geometry/topology.}
    \label{fig:diversity}
\end{figure}

Given partial bounding boxes, our model can generate a diverse set of CAD models with complete geometry and topology. Inspired by RePaint~\cite{lugmayr2022repaint}, we replace a subset of tokens with the provided bounding boxes, which are diffused to the corresponding time step during the bounding box generation stage. To enhance result quality, we adopt the same resampling technique as RePaint. After that, the generation of UDFs and SDF follows the same process as unconditional generation. Figure \ref{fig:diversity} illustrates the diversity of results produced from different Gaussian noise inputs. 

\subsection{Constructive BR-DF Geometry}
BR-DF is a set of 3D scalar fields. Constructive solid geometry operations—such as intersection and union—become applicable to B-rep models through their BR-DF representations. Figure~\ref{fig:union} illustrates such examples. Animations of more examples are provided in the supplementary video.

Suppose we are to take intersection/union of multiple B-Rep models, each of which is a surface SDF and an arbitrary number of face UDFs. The surface SDF of a combined model follows the standard formula (i.e., max for intersection and min for union). Computing face UDFs of the combined model is not trivial. Since we only care the order of UDF values (i.e., minimum face or co-minimum curve/point), we use each UDF without modification.


\begin{figure*}[t]
    \centering
    \includegraphics[width=\linewidth]{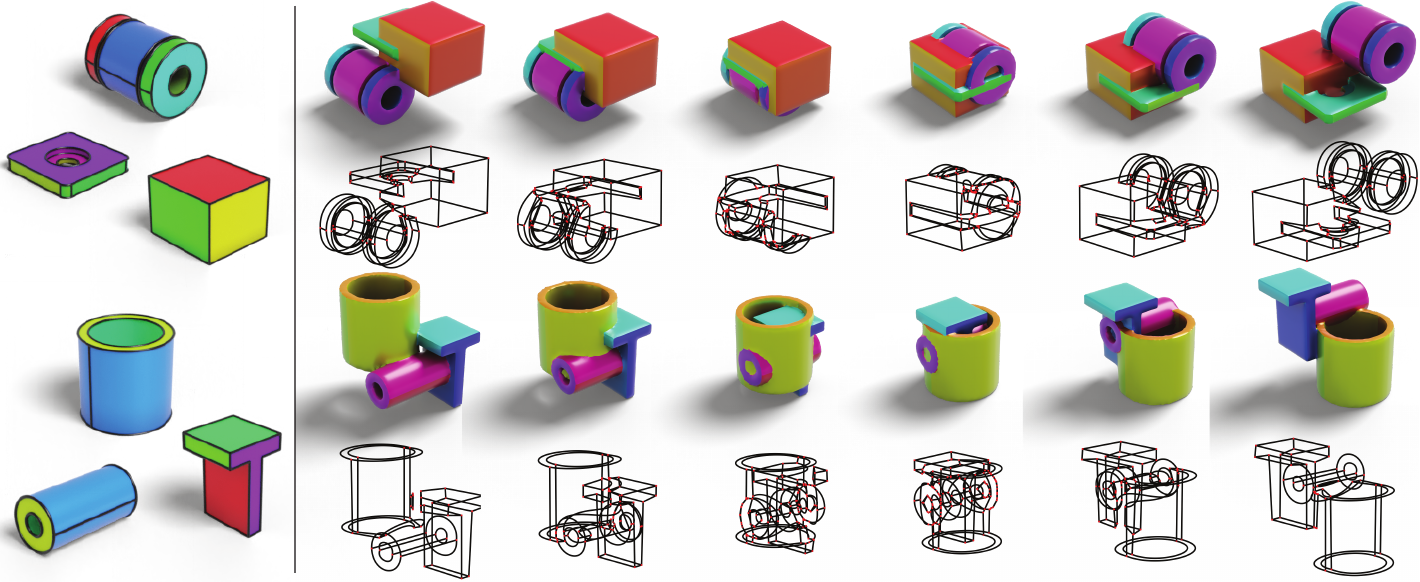} 
    \caption{Constructive BR-DF Geometry. Multiple B-Rep models (left) are converted to their BR-DF representations and combined. The resulting BR-DF representation is then converted back to a valid B-Rep model using the MCT algorithm.}
    \label{fig:union}
\end{figure*}

\begin{figure}[h!]
    \centering
    \includegraphics[width=\linewidth]{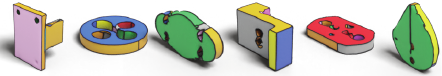} 
    \caption{Failure examples occur at thin geometries and where irregular meshes are generated by the marching cubes.}
    \label{fig:failure}
\end{figure}

\subsection{Ablation Studies}
To evaluate the effectiveness of the two-stage generation process, we conduct an ablation study by removing the bounding box generation module and using only the BR-DF Latent Diffusion Module, as described in Sec.~\ref{sec:latent_diffusion}. Since the number of faces forming a solid varies and is unknown during inference, we randomly duplicate faces during training until the maximum face count is reached. At inference, we apply non-maximum suppression (NMS) with a predefined threshold to eliminate duplicate face latents. Table~\ref{table:ablation_gt} presents the results trained on the DeepCAD dataset. While the CAD metrics (Novel, Unique, Valid) remain similar, the distribution metrics (COV, MMD, JSD) show a significant gap compared to the full two-stage model. This suggests that joint denoising for a variable number of latents is considerably more challenging for high-dimensional features (face latents) than for low-dimensional features (bounding box coordinates). Additionally, we report results using ground-truth bounding boxes.

\subsection{Failure Cases}
Figure \ref{fig:failure} presents failure cases and show the limitations of our method. Note that these failures are not caused by the Marching Cubes and Triangles (MCT) algorithm but rather by defects in the BR-DF generative model. MCT is guaranteed to convert BR-DF with a 100\% success rate. Improving the generative model will significantly mitigate these issues.


\begin{table}[t]
    \centering
    \begin{tabular}{l|ccc}
        \toprule
        Method & CD$\downarrow$ & IR$\downarrow$ & STR$\uparrow$  \\
        \midrule
        DeepCAD & $7.6\text{e-4}$ & 3.21\% & 95.21\% \\
        $\text{Ours}_\text{raw}$ & $6.6\text{e-4}$/$6.9\text{e-4}$ & 0.00\% & 100\%\\
        $\text{Ours}_\text{latent}$ & $7.5\text{e-4}$/$8.0\text{e-4}$ & 0.00\% & 93.33\% \\
        \bottomrule
    \end{tabular}
    \caption{Reconstruction evaluation. For $\text{Ours}_\text{raw}$ and $\text{Ours}_\text{latent}$, we report results with and without post-processing for CD.}
    \label{table:recon_eval}
    \vspace{-0.2cm}
\end{table}

{
\setlength{\tabcolsep}{3pt}
\begin{table}[t]
    \centering
    \begin{tabular}{lccc|ccc}
        \toprule
        Bbox & COV & MMD & JSD & Novel & Unique & Valid\\
        & $\%\uparrow$ & $\downarrow$ & $\downarrow$ & $\%\uparrow$ & $\%\uparrow$ & $\%\uparrow$\\
        \midrule
        Pred & 73.7 & 1.09 & 1.40 & 100. & 99.1.& 100.*\\
        w/o & 66.1 & 1.29 & 1.52 & 99.9 & 99.3. & 100.* \\
        GT &  75.8 &  1.06 & 1.22 & 99.8 & 98.7 &  100.*\\
        \bottomrule
    \end{tabular}
    \caption{Effectiveness of the two-stage generation process. Ablation using predicted bounding box (pred), no bounding box (w/o), and ground-truth bounding box (GT).}
    \label{table:ablation_gt}
\end{table}
}

\section{Conclusions}
We introduced B-Rep Distance Functions, a novel geometric representation for CAD B-Rep models using volumetric distance functions. BR-DF encodes surface geometry as an SDF and represents B-Rep elements (vertices, curves, faces) and their connectivity as per-face UDFs. Our Marching Cubes and Triangles algorithm always converts any BR-DF into a watertight faceted B-Rep model.
Compared to other B-Rep modeling methods, BR-DF is 3D scalar fields, making its generation immediately applicable to popular approaches such as diffusion models.
Future work is to enhance generation quality and handle more complex models.

\section{Acknowledgements}
This research is partially supported by NSERC Discovery Grants, NSERC Alliance Grants, and John R. Evans Leaders Fund (JELF). We thank the Digital Research Alliance of Canada and BC DRI Group for providing computational resources.

{
    \small
    \bibliographystyle{ieeenat_fullname}
    \bibliography{main}
}

\clearpage
\setcounter{page}{1}
\maketitlesupplementary

\subsection{Extracting boundary segments}
\label{sec:post1}
\begin{figure}[ht]
\centering
\includegraphics[width=\linewidth]{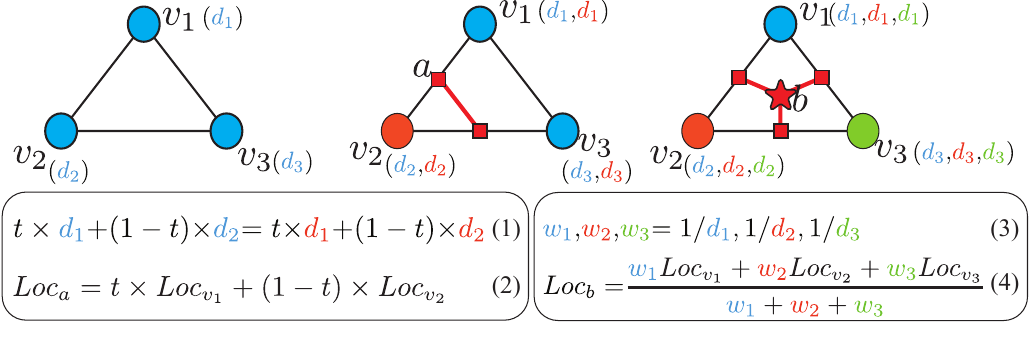}
\caption{Brep Vertex/Curve Interpolation.}
\label{fig:supp_inter}
\end{figure}

The Marching Cubes and Triangles (MCT) algorithm traces all the mesh triangles and extracts the boundaries following 3-way rules (1) If the minimal faces of all the three vertices are the same, no point or curves are created; (2) If two vertices share the same minimal face, a B-Rep curve is created to cut the triangle via linear interpolation; (3) If all three minimal faces are different, a B-Rep vertex and three incident B-Rep curves are created. The equations for computing the B-Rep points on the mesh edges or inside the triangles are shown in Figure~\ref{fig:supp_inter}.

The boundary between every two neighbor faces is defined by a set of intersection points (marked as red cubes or stars in Figure~\ref{fig:supp_inter}) and their connections. The boundary is guaranteed to begin and end with 3-degree junctions (red stars). 

\subsection{Boundary Smoothing}
\label{sec:post2}
For the boundaries of every two neighbor faces, we fit them to either a straight line (1-degree polynomial function) or a curve line (3-degree polynomial function). The start and end points are preserved by assigning them a high fitting weight. After fitting, the adjusted boundary curves are projected back onto the mesh to update the positions of the intersection points.

\subsection{Establishing Faces and Boundaries}
\label{sec:post3}
\begin{figure}[ht]
\centering
\includegraphics[width=\linewidth]{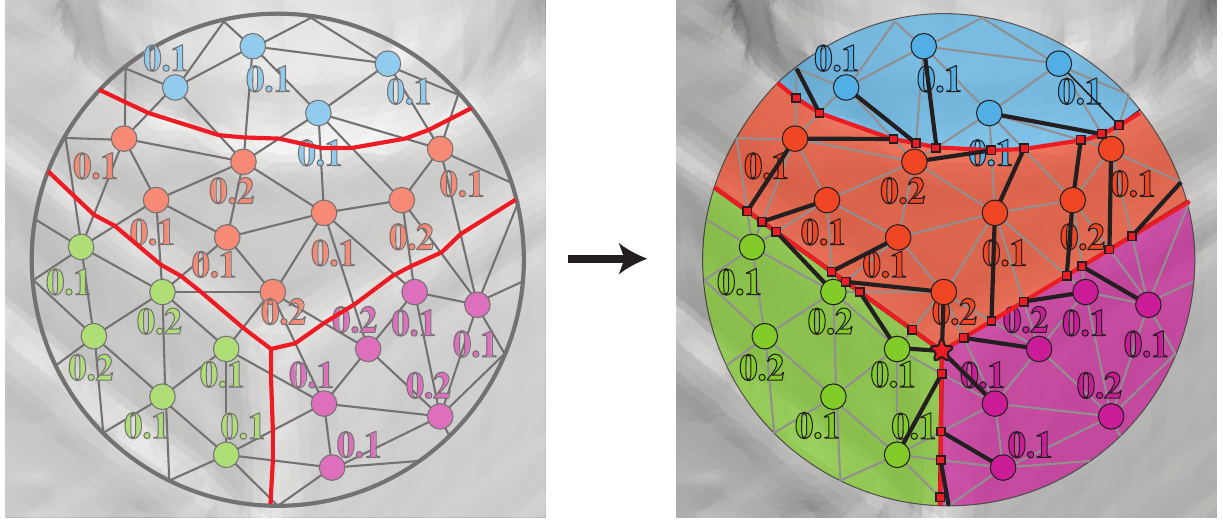}
\caption{Illustration of face and boundary establishment.}
\label{fig:supp_line}
\end{figure}

To properly define faces and boundaries, new triangles must be introduced within the mesh so that boundary curves become actual mesh edges. Figure~\ref{fig:supp_line} illustrates this process, where dark lines represent newly added edges, and different colors indicate separated faces.

\subsection{Mesh Smoothing with Topology Preservation}
\label{sec:post4}

The initial mesh generated by Marching Cubes often has rough surfaces and rounded boundary regions. To improve quality, we apply multiple iterations of Laplacian smoothing and Quadric-based edge-collapse remeshing (with the PyMeshLab package). During this process, boundary vertices are averaged to maintain geometric consistency while preserving the mesh’s overall structure.

\section{Additional Results}
\begin{figure*}[ht]
\centering
\includegraphics[width=\linewidth]{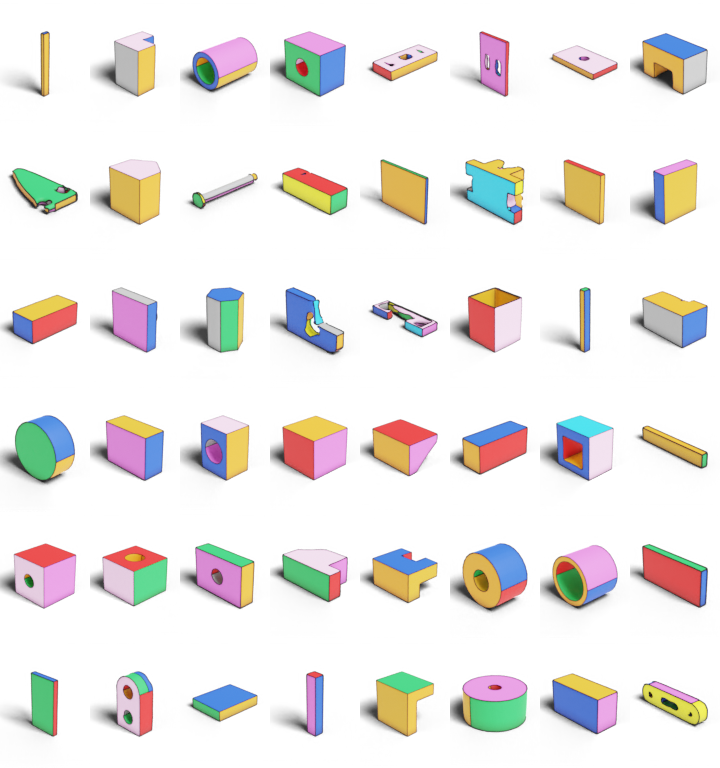}
\caption{Additional Results.}
\label{fig:supp_result1}
\end{figure*}

\begin{figure*}[ht]
\centering
\includegraphics[width=\linewidth]{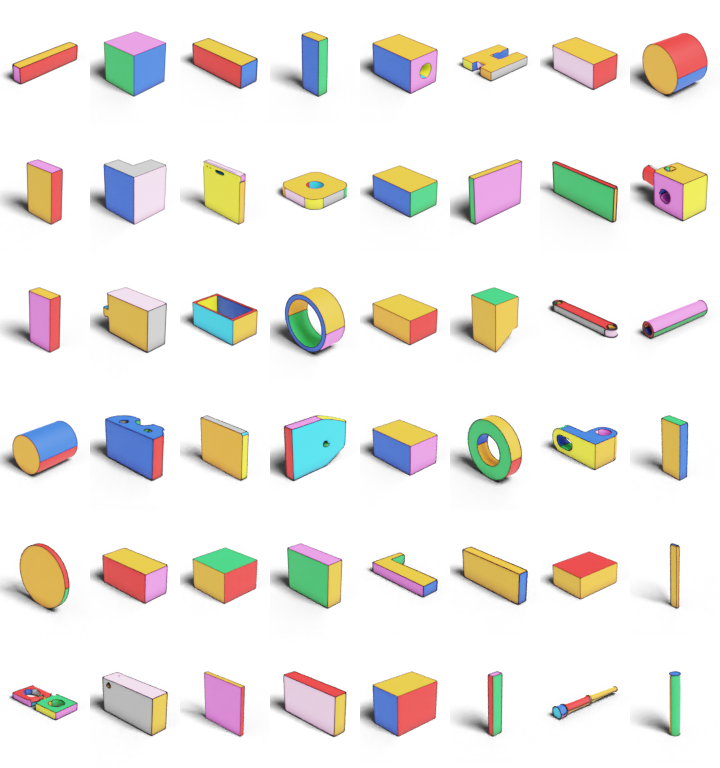}
\caption{Additional Results.}
\label{fig:supp_result2}
\end{figure*}

\begin{figure*}[ht]
\centering
\includegraphics[width=\linewidth]{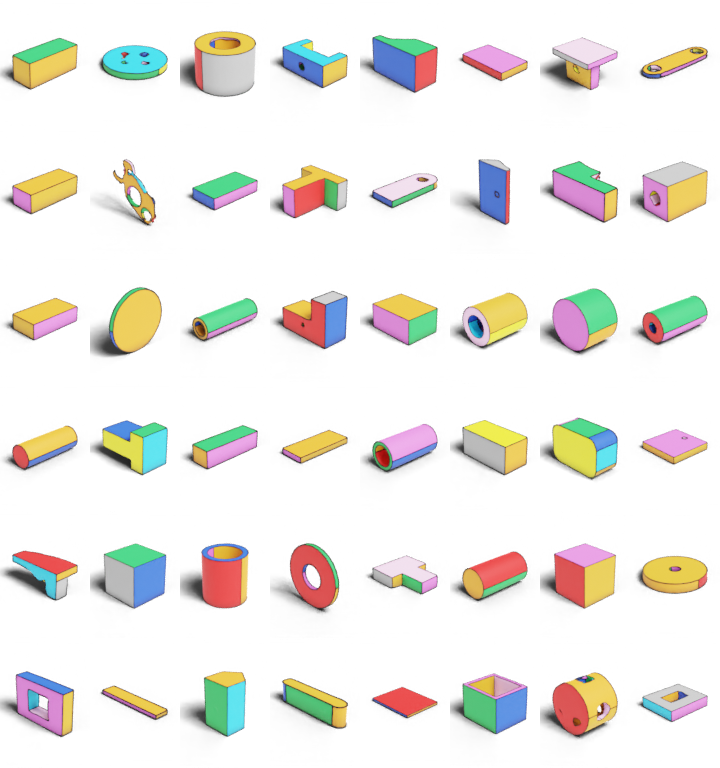}
\caption{Additional Results.}
\label{fig:supp_result3}
\end{figure*}

\begin{figure*}[ht]
\centering
\includegraphics[width=\linewidth]{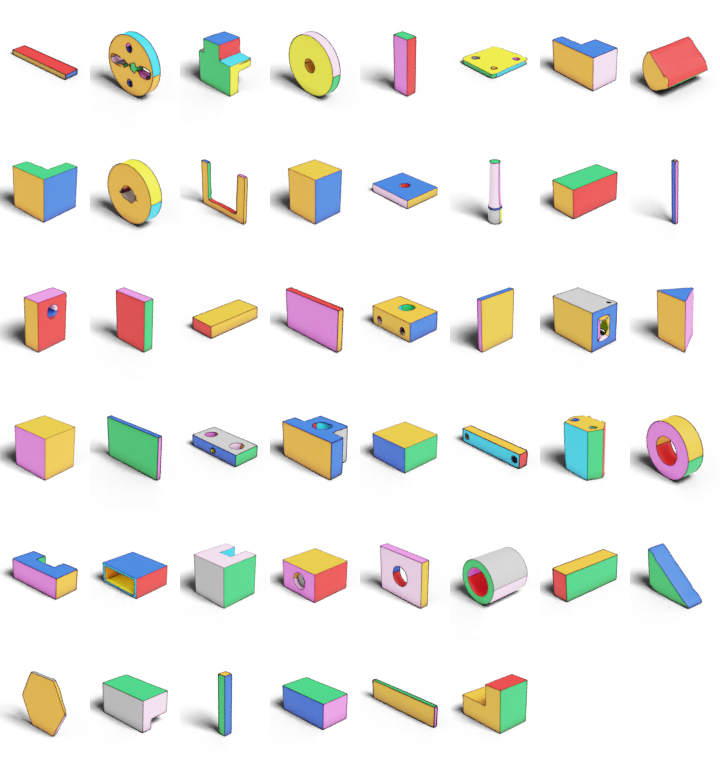}
\caption{Additional Results.}
\label{fig:supp_result4}
\end{figure*}

Figures~\ref{fig:supp_result1},\ref{fig:supp_result2},\ref{fig:supp_result3},\ref{fig:supp_result4} shows additional results on DeepCAD dataset.

\end{document}